\documentclass[10pt,twocolumn,letterpaper]{article}

\usepackage{cvpr}
\renewcommand\footnotemark{}

\definecolor{turquoise}{cmyk}{0.65,0,0.1,0.3}
\definecolor{purple}{rgb}{0.65,0,0.65}
\definecolor{dark_green}{rgb}{0, 0.5, 0}
\definecolor{orange}{rgb}{1.0, 0.65, 0.2}
\definecolor{red}{rgb}{0.8, 0.2, 0.2}
\definecolor{darkred}{rgb}{0.6, 0.1, 0.05}
\definecolor{blueish}{rgb}{0.0, 0.3, .6}
\definecolor{light_gray}{rgb}{0.7, 0.7, .7}
\definecolor{pink}{rgb}{0.9, 0, 0.6}
\definecolor{greyblue}{rgb}{0.25, 0.25, 1}
\definecolor{teal}{rgb}{0.0, 0.4, 0.4}

\def \customparskip {.5em}
\renewcommand{\paragraph}[1]{\vspace{\customparskip}\noindent\textbf{#1}}

\definecolor{cvprblue}{rgb}{0.21,0.49,0.74}
\usepackage[pagebackref,breaklinks,colorlinks,allcolors=cvprblue]{hyperref}

\def\paperID{17741} 
\def\confName{CVPR}
\def\confYear{2025}

\title{Improving Generative Pre-Training: An In-depth Study\\of Masked Image Modeling and Denoising Models\thanks{$^{\dagger}$ Corresponding author: dbmin@ewha.ac.kr}}

\author{Hyesong Choi\textsuperscript{\rm 1},\;
Daeun Kim\textsuperscript{\rm 1},\;
Sungmin Cha\textsuperscript{\rm 2},\;
Kwang Moo Yi\textsuperscript{\rm 3},\;
Dongbo Min\textsuperscript{\rm 1}$^{\dagger}$\\
\textsuperscript{\rm 1}Ewha W. University\;\;\;
\textsuperscript{\rm 2}New York University\;\;\;
\textsuperscript{\rm 3}University of British Columbia\\
}

\begin{document}
\maketitle

\begin{abstract}
In this work, we dive deep into the impact of additive noise in pre-training deep networks.
While various methods have attempted to use additive noise inspired by the success of latent denoising diffusion models,
when used in combination with masked image modeling, their gains have been marginal when it comes to recognition tasks.
We thus investigate why this would be the case, in an attempt to find effective ways to combine the two ideas.
Specifically, we find three critical conditions: corruption and restoration must be applied within the encoder, noise must be introduced in the feature space, and an explicit disentanglement between noised and masked tokens is necessary.
By implementing these findings, we demonstrate improved pre-training performance for a wide range of recognition tasks, including those that require fine-grained, high-frequency information to solve.
\end{abstract}    
\section{Introduction}
\label{sec:intro}

Foundational models~\cite{dosovitskiy2020image, liu2021swin} have appeared as effective solutions to various problems~\cite{carion2020end, ranftl2021vision, zhu2020deformable, zheng2021rethinking, chen2021pre}.
However, 
these models are immense and notoriously data-hungry, requiring vast amounts of labeled data to reach their full potential.
For example, Stable Diffusion~\cite{rombach2022high} is trained on 400 million images from the  LAION-400m dataset~\cite{schuhmann2021laion}, and CLIP~\cite{radford2021learning} also with 400 million images.
Naturally, to alleviate this data constraint,
pre-training of these models that are task-agnostic 
based on self-supervision~\cite{chen2020simple, he2020momentum, grill2020bootstrap, bao2021beit, he2022masked, xie2022simmim} has drawn interest from the research community.

\begin{figure}[t!]
	\centering
	\includegraphics[width=1\columnwidth]{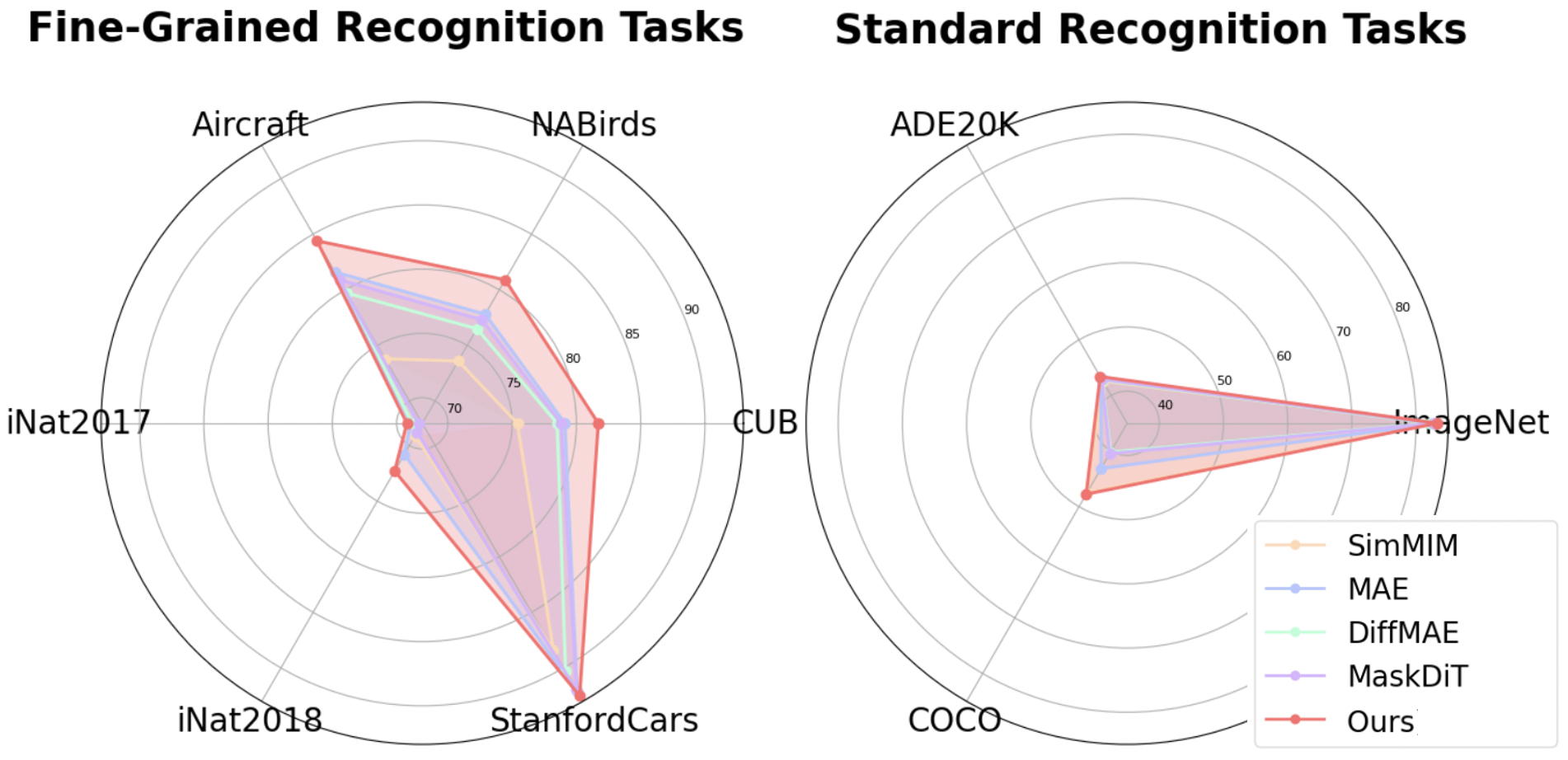}
	\caption{We find that noise-based pre-training, when applied in the right way, can enhance transfer learning ability. Leveraging the insights, we introduce a novel pre-training setup combining masking and noising, outperforming MIM baselines~\cite{he2022masked, xie2022simmim} and recent noise-based generative approaches~\cite{wei2023diffusion, zheng2023fast} across a wide range of recognition tasks, including fine-grained recognition.
	}
	\label{fig:teaser}
	\vspace{-0.2cm}
\end{figure}
\begin{figure*}[t!]
	\centering
	\includegraphics[width=2.1\columnwidth]{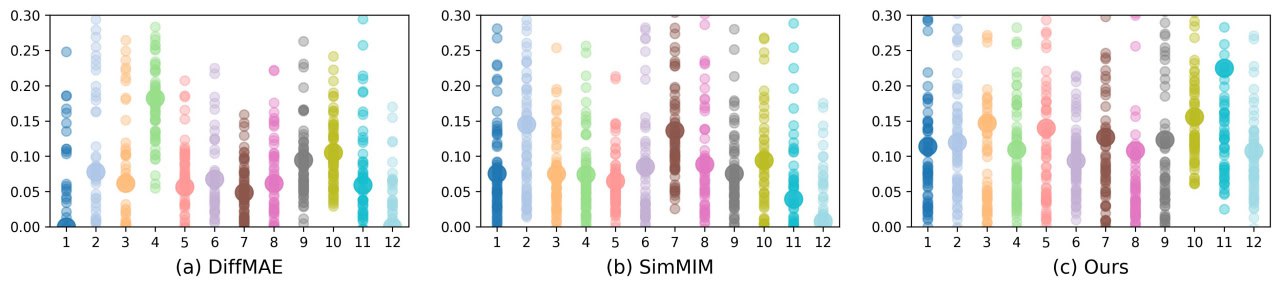}
	\vspace{-2em}
	\caption{We display the KL divergence among attention distributions across different heads (indicated by small dots) and the mean KL divergence (represented by large dots) in each layer for (a) a recent generative model~\cite{wei2023diffusion}, (b) a representative masked image model~\cite{xie2022simmim}, and (c) our method. This assesses whether various attention heads capture diverse frequency information, where higher KL divergence indicates broader frequency capture. Our method demonstrates a greater capacity for capturing diverse frequency information than MIM and generative approaches, which explains why it performs well across a wide range of recognition tasks, including fine-grained tasks.}
	\label{fig:kl_divergence}
\end{figure*}
\begin{figure*}[t!]
	\centering
	\includegraphics[width=2\columnwidth]{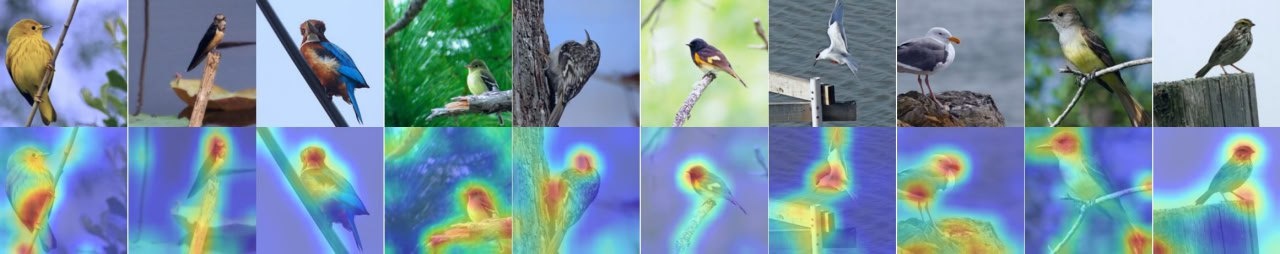}
	\caption{We visualized the self-attention maps for the image classification token in the final layer of our model on a fine-grained visual categorization benchmark. The proposed method captures a range of frequencies by focusing effectively on both key features and fine details within complex scenes.}
	\label{fig:qualitative}
	\vspace{-0.2cm}
\end{figure*}

A popular trend is based on masked image modeling (MIM)~\cite{bao2021beit, he2022masked, xie2022simmim}, thanks to their simplicity and effectiveness.
These models mask parts of images and learn feature representations by learning to reconstruct them.
While effective, these masked image models primarily capture low-frequency semantics of the scene~\cite{he2022masked}, with the masking and reconstruction process encouraging the model to focus on the general structure and contextual patterns. As a result, they perform poorly on tasks that require high-frequency details of images, such as fine-grained recognition, as shown in \cref{fig:teaser} and 
later in our experiments.

In other branches of computer vision~\cite{rombach2022high,dalle,imagen}, latent denoising diffusion models have shown to be effective in capturing high-frequency details.
These approaches, in particular, excel at generating images and videos with high fidelity.
This 
demonstrates that the latent representations employed by these models must contain high-frequency textures they generate---a potential for downstream tasks that demand fine-grained details~\cite{cub, nabirds, inat2017, inat2018, stanfordcars, aircraft}.

Naturally, there have been attempts to utilize the potential of denoising models, by incorporating their `noising'-based training to pre-training of deep features.
That is, learning to denoise similarly to how masked image models learn to recover images, or even learning to perform both together.
For example, Diffused Masking strategy~\cite{wei2023diffusion} uses tokens with noise added alongside clean ones, replacing the conventional strategy of masking tokens.
The hybrid Masking~\cite{zheng2023fast} combines masking and noising, by utilizing both masked tokens and noisy visible tokens within its framework.
Both of them, however, do not provide notable gains over conventional masking-based approaches to recognition benchmarks~\cite{imagenet, ade20k, coco}, as well as in fine-grained recognition~\cite{cub, nabirds, inat2017, inat2018, stanfordcars, aircraft}; see \cref{fig:teaser}.

In this paper, we dive deep into why this framework shows limited gains in recognition tasks and discover that it is possible to benefit from noise-based pre-training, much like the denoising diffusion models, \textit{if done in the right way}. By doing so, we redefine the role of generative models within self-supervised representation learning, revisiting the term `generative pre-training' to establish a focused domain of study. In more detail, we find the following: (1) \textbf{corruption and restoration must be applied within the encoder} while training, unlike existing design choices~\cite{wei2023diffusion, zheng2023fast}, as the encoder is what is ultimately transferred to extract features in downstream tasks; (2) \textbf{noise must be added at the feature level}, and is particularly effective when added at lower layers of the encoder, which is where high-frequency details are present; and (3) using both the masking strategy and the noising strategy can interfere with each other and \textbf{these strategies must be explicitly disentangled}---we ensure this by suppressing the attention between the two different types of tokens.

With these findings, we design a novel pre-training setup that effectively utilizes both masking and noising. 
Our approach captures a broader range of frequency information as shown in \cref{fig:kl_divergence} and \cref{fig:qualitative}, which enhances transferability across a variety of downstream tasks and benchmarks---CUB-200-2011~\cite{cub}, NABirds~\cite{nabirds}, iNaturalist
2017~\cite{inat2017}, iNaturalist 2018~\cite{inat2018}, Stanford Cars~\cite{stanfordcars}, Aircraft~\cite{aircraft}, ImageNet~\cite{imagenet}, ADE20K~\cite{ade20k}, and COCO~\cite{coco}---achieving up to an 8.1\% gain over MIM methods and an 8.0\% improvement over recent generative baselines; see \cref{fig:teaser}.

To summarize, our contributions are:
\begin{itemize}
    \item we provide a thorough empirical study on why current noising-based pre-training approaches~\cite{wei2023diffusion, zheng2023fast} do not provide noticeable gains for recognition tasks;
    \item we provide guidelines from our detailed study on how to use noise within pre-training;
    \item and with our findings, we propose a novel pre-training method that outperforms the state-of-the-art on a wide range of recognition tasks, including fine-grained tasks.
\end{itemize}

\section{Preliminary and Related Works}

As the intuitions and the findings behind our work are 
grounded on masked image modeling (MIM) and denoising diffusion models, we first review them for completeness.

\subsection{Masked Image Modeling}
\label{sec2.1}

The core idea behind MIM is to randomly mask portions of images (tokens) and then learn to reconstruct them in a self-supervised manner.

\paragraph{Random masking.} 
Formally, let $X \in \mathbb{R}^{N \times L \times D}$ represent an image sequence (tokens), where $N$ denotes the batch size, $L$ is the number of tokens in each image, and $D$ represents the dimension of each token.
Let us further denote the mask generation process as $M=\Phi_{\text{M}}(X, \gamma)$, where $\gamma$ is the masking ratio, and $M \in \{0,1\}^{N \times L}$ is the mask that is generated.
We can then write the remaining visible tokens as
$X_{\text{masked}} = M \odot X$, where $\odot$ is the Hadamard product.

\paragraph{Reconstruction.} 
To learn to reconstruct the original tokens $X$ back from only the visible ones $X_{\text{vis}}$, training of MIMs typically relies on mean squared error (MSE).
With the token predictions $\hat{X}$ from MIM framework that takes as input the masked visible tokens $X_{\text{masked}}$,
we minimize the loss: 
\begin{equation}
    \mathcal{L}_{\text{MIM}} = \frac{1}{\sum_{k,l} M_{k,l}} \sum_{k=1}^{N} \sum_{l=1}^{L} M_{k,l} \Vert \hat{X}_{k,l} - \bar{X}_{k,l} \Vert^2,
\end{equation}
where $\bar{X}=X_\text{vis}$.

\paragraph{Recent works.}
MIM approaches~\cite{he2022masked, xie2022simmim, choi2024emerging, choi2025salience, bao2021beit, yi2022masked, dong2022bootstrapped, chen2024context} adapt the concept of Masked Language Modeling (MLM) from NLP. 
BEiT~\cite{bao2021beit} applies MLM-like pre-training to images using discrete visual tokens generated by a pre-trained dVAE. MAE~\cite{he2022masked} focuses only on visible patches in the encoder, predicting masked pixel values through a decoder. SimMIM~\cite{xie2022simmim} uses both visible and masked patches in the encoder and predicts original pixels directly. 
Recent advances~\cite{choi2024emerging, choi2025salience} focus on masked tokens for fast convergence and performance improvement.

\subsection{Denoising Diffusion Model}

Denoising diffusion models are trained by progressively corrupting the input data with additive Gaussian noise, which is then a denoiser is trained to recover the data.
This is in a similar spirit to MIMs, but unlike MIMs, the theoretical foundations allow the model to be used to generate completely new data, hence they are generative.~\cite{song2020denoising}
Various works~\cite{rombach2022high, hedlin2024unsupervised, hedlin2024keypoint, luo2024diffusion} have thus sought to utilize diffusion models beyond purely generative applications.

\paragraph{Forward diffusion.} 
Forward diffusion 
iteratively adds noise to an input image sequence $X \in \mathbb{R}^{N \times L \times D}$ over $T$ time steps. 
At each time step $t$, a noise schedule $\beta_t \in \mathbb{R}$ controls the amount of noise added, where $\beta^t$ is a scalar that determines the noise level at time step $t$. 
The corrupted representation $X^t$ at step $t$ is then defined as:
\begin{equation} 
    X^t = \sqrt{1 - \beta^t} \cdot X^{t-1} + \sqrt{\beta^t} \cdot \epsilon, 
\end{equation}
where $\epsilon \sim \mathcal{N}(\mathbf{0}, \mathbf{I})$ is the Gaussian noise with a zero matrix $\mathbf{0}$ and an identity covariance matrix $\mathbf{I}$. 
This iterative process gradually \emph{diffuses} the data towards Gaussian noise as $t$ approaches $T$.

\paragraph{Denoising.} 
With the corrupted signal, the denoiser then learns to undo this corruption, effectively allowing the model to traverse back through the diffusion process, that is, transform Gaussian noise to clean data that follows the data distribution.
Specifically,
starting from $X^T$, the model learns to predict the clean image $X^0$ by estimating the intermediate states through a denoising function $\Phi_{\text{denoise}}$, which is typically parameterized as a neural network. 
The denoising step at time $t$ can be represented as:
\begin{equation} 
    \hat{X}^{t-1} = \Phi_{\text{denoise}}(X^t, t),
\end{equation}
where $\hat{X}^{t-1}$ represents the denoised estimate at time $t-1$. 
Training of $\Phi_{\text{denoise}}(X^t, t)$ is performed through various variations of the original DDIM~\cite{song2020denoising} and DDPM~\cite{ho2020denoising} methods, including recent family of Rectified Flow models~\cite{liu2022flow}, but are all essentially focusing on obtaining $\hat{X}^{t-1}$ estimates in some form that will accurately lead toward $X^0$ through various solvers~\cite{lu2022dpm, karras2022elucidating}.

\paragraph{Recent works.}
Denoising diffusion models~\cite{ho2020denoising, nichol2021improved, rombach2022high, dalle, imagen} have gained prominence in generative tasks for their ability to produce high-quality, detailed images.
DDPM~\cite{ho2020denoising} introduced the foundational framework, where  Gaussian noise is gradually added to an image and then removed in a reverse process, Improved DDPM~\cite{nichol2021improved} enhanced this approach with modifications in noise scheduling and model architecture. LDM~\cite{rombach2022high} further improved efficiency by operating in a compressed latent space rather than pixel space, allowing various practical applications.

\subsection{Pre-training via denoising}


With preliminaries on MIMs and denoising diffusion models, we now review two representative works that aim to marry the two schools of thought into a single pre-training framework.

\begin{figure}[t!]
	\centering
	\includegraphics[width=0.6\columnwidth]{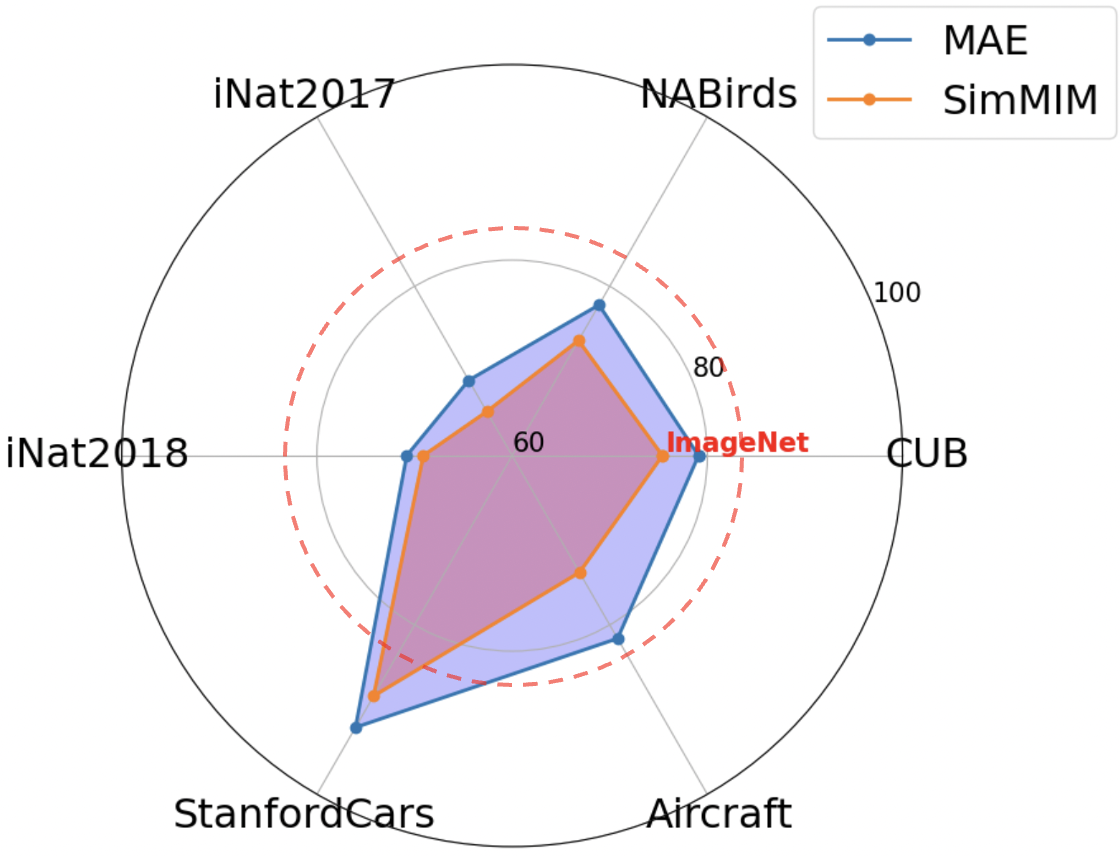}
	\caption{Fine-grained visual categorization (FGVC) is a critical benchmark for evaluating recognition models 
    and require detailed, localized feature learning. 
    However, MIM approaches~\cite{he2022masked, xie2022simmim} show limitations on FGVC tasks, with the radar graph revealing substantial room for improvement to reach the ideal boundary and also to the performance on the standard recognition task.}
	\label{fig:mim_fgvc}
	\vspace{-0.2cm}
\end{figure}

\paragraph{DiffMAE.} DiffMAE~\cite{wei2023diffusion} proposes a framework combining diffusion-based generative modeling and MIM. 
DiffMAE operates by introducing noised tokens instead of masking tokens out.
Rather than 
utilizing a masked token, the
noise is applied to the regions indicated by binary mask $M \in \{0,1\}^{N \times L}$.
Following the noise schedule defined by $\alpha_t \in \mathbb{R}$, 
Gaussian noise $\epsilon \sim \mathcal{N}(\mathbf{0}, \mathbf{I})$  is applied at each step of the diffusion process.
We thus write the noise addition (masking) process as:
\begin{equation}
    X \odot (1-M) + \left(\sqrt{\alpha_t} \cdot X + \sqrt{1 - \alpha_t} \cdot \epsilon \right) \odot M
    ,
    \label{eq:diffmae}
\end{equation}
In what follows, for ease in notation, we denote the visible image tokens in $X_{\text{masked}}$, that is, tokens in $X \odot M$ as $x_v\in \mathrm{R}^{N\times D}$ and the noisy tokens, the tokens in $\left(\sqrt{\alpha_t} \cdot X + \sqrt{1 - \alpha_t} \cdot \epsilon \right) \odot (1-M)$ as $x_n\in \mathrm{R}^{N\times D}$.
%
The model then reconstructs $X$ from the corrupted image $X_{\text{masked}}$ by denoising the noisy tokens $x_n$ through a reverse process that iteratively refines $x_n$ using visible tokens $x_v$:
\begin{equation} 
    \hat{x}_n^{t-1} = \Phi_{\text{denoise}}(x_n^t, x_v, t), 
\label{eq6}\end{equation}
where $\Phi_{\text{denoise}}$ is the denoising function.


\paragraph{MaskDiT.} MaskDiT~\cite{zheng2023fast} proposes a generative pre-training approach that leverages masked transformers for faster training of diffusion models. 
Unlike DiffMAE, MaskDiT generates a masked input $X_{\text{masked}}$, using both masked tokens and noised tokens. 
We describe the noise addition and masking process as:
\begin{equation} 
    \theta \odot M + \left( \sqrt{\alpha_t} \cdot X + \sqrt{1 - \alpha_t} \cdot \epsilon \right) \odot (1 - M),
    \label{eq:maskdit}
\end{equation}
where $\theta$ is a parameter for masked tokens, $x_m{\in \mathrm{R}^{N\times D}}$.
Similarly, as before, we denote the masked tokens $\theta \odot M$ of $X_{\text{masked}}$ as $x_m$.
Then,
the model reconstructs the noise tokens $x_n^t\in \mathrm{R}^{N\times D}$ and masked tokens $x_m$ through a function $\Phi_{\text{denoise}}$ and $\Phi_{\text{reconstruct}}$:
\begin{equation} 
    \hat{x}_m^{t-1} = \Phi_{\text{denoise}}(x_n^t, x_v, t) \quad
\hat{x} = \Phi_{\text{reconstruct}}(x_m, x_v)
.
\label{eq8}\end{equation}

\begin{figure*}[t!]
	\centering
	\includegraphics[width=1.7\columnwidth]{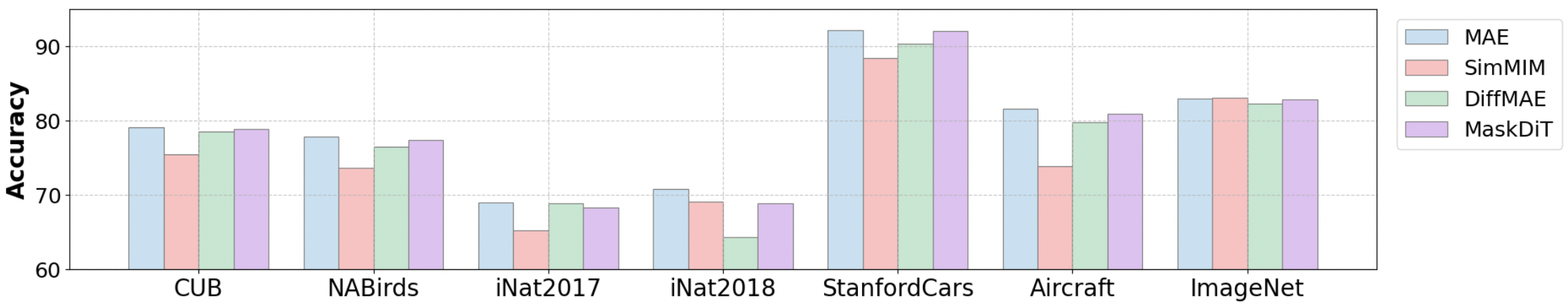}
	\caption{Our evaluations reveal that recent generative pre-training approaches~\cite{wei2023diffusion, zheng2023fast} yield limited gains over MIM baselines~\cite{xie2022simmim, he2022masked} on recognition tasks, suggesting that simply adding denoising to MIM pre-training does not inherently elevate the representation quality essential for precise recognition. 
    Except for DiffMAE~\cite{wei2023diffusion}, we rely on the official implementation---for DiffMAE, we carefully reimplemented the method based on the manuscript as no code is available.
    For reproducibility, all implemented code has been included in the Supplementary Material.}
	\label{fig:genmim_fgvc}
	\vspace{-0.3cm}
\end{figure*}

\section{A Deep Dive into Pre-training Methods}

Given the preliminaries, unfortunately, none of the existing works show a clear benefit in combining denoising models with MIM, especially for fine-grained recognition tasks as demonstrated 
in \Cref{fig:teaser}.
Before we discuss our method, to motivate our design choices, we investigate the limitations of existing methods and 
where their potential pitfalls are.

\subsection{Limitations of Conventional Methods}

We first examine the status of Masked Image Modeling then Denoising-based Pre-training.

\paragraph{Limitations of Masked Image Modeling in Fine-Grained Recognition.}
Masked Image Modeling (MIM) initially demonstrated strong transferability in downstream tasks like image classification, positioning it as a promising approach for self-supervised learning.
Meanwhile, fine-grained recognition tasks, such as fine-grained visual categorization (FGVC), have become essential benchmarks for recognition models in computer vision due to their need for detailed, localized feature learning to distinguish between highly similar categories.
MIM approaches~\cite{xie2022simmim, he2022masked} reveal limitations on FGVC tasks, as shown in \Cref{fig:mim_fgvc}.

Specifically, we pre-trained each MIM method under identical settings and then fine-tuned them on each fine-grained datasets, such as CUB-200-2011~\cite{cub},  NABirds~\cite{nabirds},  iNaturalist
2017~\cite{inat2017}, iNaturalist 2018~\cite{inat2018}, Stanford Cars~\cite{stanfordcars}, and Aircraft~\cite{aircraft} to obtain the transferred performance. 
As observed in the radar graph, MIM pre-training alone falls short of achieving high accuracy, and the remaining gap compared to the ideal boundary and the ImageNet baseline (performance on the standard recognition task), highlights significant room for improvement in the FGVC task.
One possible reason is that the reconstruction objective MIM relatively focuses on broader, lower-frequency structures as illustrated in \Cref{fig:kl_divergence}, which may limit its ability to capture the fine-grained details that are essential to distinguishing 
across visually similar categories.


\begin{figure}[t!]
	\centering
	\includegraphics[width=0.6\columnwidth]{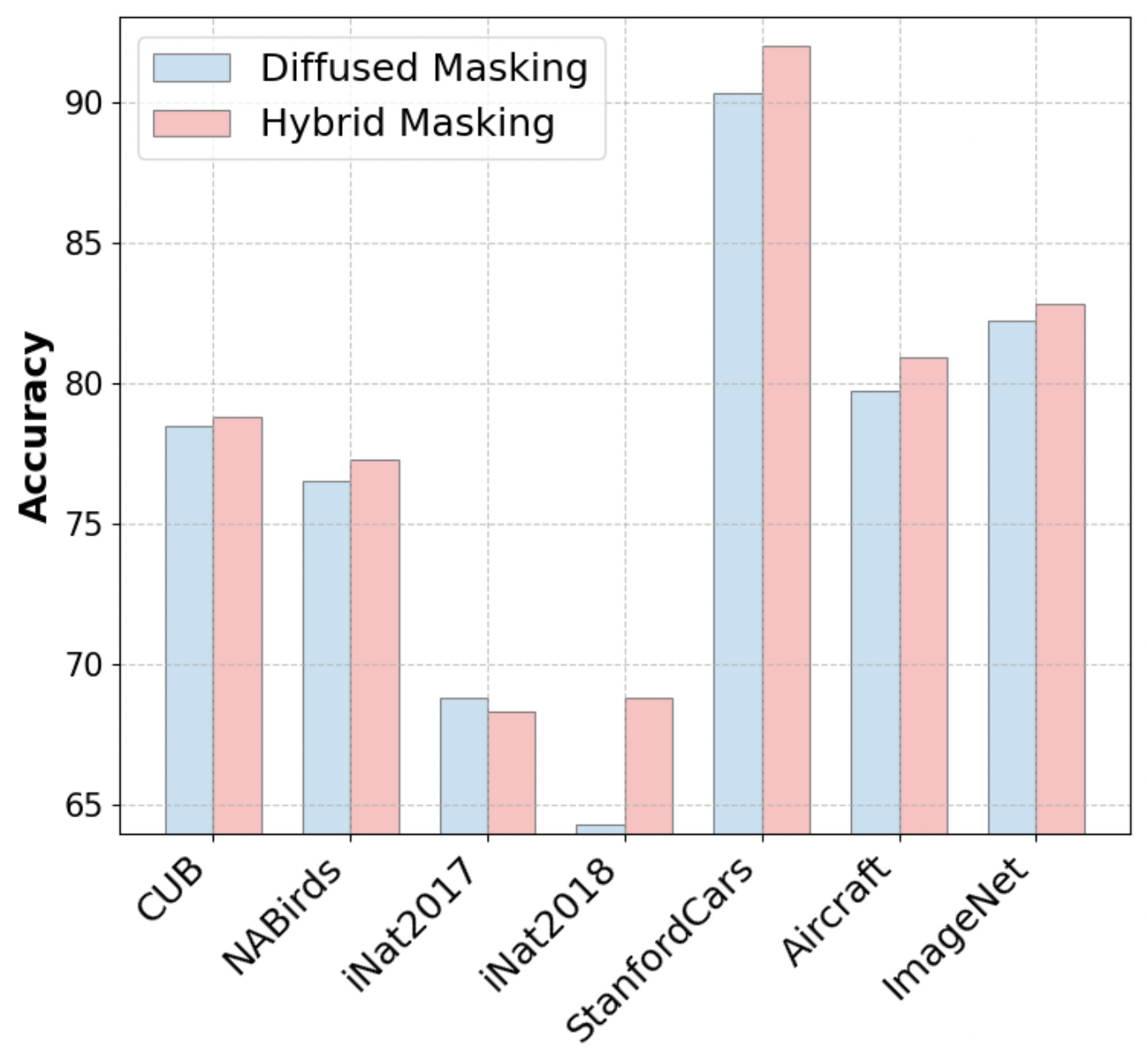}
	\vspace{-1em}
	\caption{Hybrid masking achieved slightly better performance across datasets, as it consistently retains a fully masked portion that enhances semantic discriminability. In contrast, diffused masking relies on tokens with random noise intensities, which may limit its effectiveness.}
	\label{fig:masking_methods}
	\vspace{-0.2cm}
\end{figure}

\paragraph{Challenges of Recent Denoising-based Pre-Training for Recognition Tasks.}
Recent denoising-based pre-training approaches~\cite{wei2023diffusion, zheng2023fast}, which utilize additive noise to denoise masked image modeling, were designed to improve representation learning. 
To be precise, DiffMAE~\cite{wei2023diffusion} is designed for both recognition and generative tasks, while MaskDiT~\cite{zheng2023fast} focuses specifically on generative tasks.
However, our evaluations, along with results reported in the study~\cite{zheng2023fast}, reveal that these generative methods struggle to deliver notable gains on recognition tasks compared to representative MIM baselines~\cite{xie2022simmim, he2022masked}.

In \Cref{fig:genmim_fgvc}, we show results of pre-training with recent denoising-based methods,
DiffMAE~\cite{wei2023diffusion} and MaskDiT~\cite{zheng2023fast}, and MIM baselines, SimMIM~\cite{xie2022simmim} and MAE~\cite{he2022masked}.
For a fair comparison, all methods were pre-trained under the same settings and subsequently fine-tuned on each recognition dataset. 
Notably, recent generative pre-training methods show minimal performance differences from MIM baselines on standard recognition tasks such as ImageNet~\cite{imagenet} image classification.
It even falls short in capturing the fine-grained features essential for FGVC tasks on datasets such as CUB-200-2011~\cite{cub},  NABirds~\cite{nabirds},  iNaturalist
2017~\cite{inat2017}, iNaturalist 2018~\cite{inat2018}, Stanford Cars~\cite{stanfordcars}, and Aircraft~\cite{aircraft}, where subtle distinctions are crucial. 
Except for DiffMAE~\cite{wei2023diffusion}, we rely on the official implementation---for DiffMAE, we carefully reimplemented the method based on the manuscript.
To ensure transparency, all implemented code is provided in the Supplementary Material.
This limitation indicates that simply incorporating a denoising process into MIM pre-training does not inherently elevate the representation quality essential for precise recognition tasks.

\subsection{How should noising and masking be combined?}
\label{sec3.2}

Despite integrating MIM with denoising diffusion models, these methods~\cite{wei2023diffusion,zheng2023fast} yield no notable gains over standard MIM approaches~\cite{xie2022simmim,he2022masked} on recognition tasks.
We first focus on how noising and masking are combined in recent baselines to uncover why they fall short on recognition tasks.
Specifically, we look into the two representative cases of integrations, DiffMAE~\cite{wei2023diffusion} and MaskDiT~\cite{zheng2023fast}.
We note that these methods represent the two different ways in which noise can be added: 
\begin{itemize}
    \item at visible tokens, \ie, \textbf{Diffused Masking}~\cite{wei2023diffusion}; or
    \item at masked tokens, \ie, \textbf{Hybrid Masking}~\cite{zheng2023fast}.
\end{itemize}


\emph{Diffused masking} employs a noisy, diffused token alongside a clean visible token, as specified in (\ref{eq:diffmae}).
On the other hand, \emph{hybrid masking} utilizes both a masked token and a noisy token, as described in (\ref{eq:maskdit}).
As outlined in Section~\ref{sec2.1}, the conventional MIM setup consists of both masked and visible tokens; simply put, diffused masking modifies this by replacing the masked token with a noisy token, while hybrid masking keeps the masked token but adds noise to the visible token.
Thus, diffused masking focuses only on denoising, as in (\ref{eq6}), while hybrid masking simultaneously performs noised token denoising and masked token reconstruction, as in (\ref{eq8}).

We evaluated these two masking methods used by recent baselines~\cite{wei2023diffusion, zheng2023fast} to determine which approach performs better across a wide range of recognition tasks.
For a fair comparison, we re-implemented both methods to ensure all conditions other than the masking method—such as model architecture and training schedule—are identical.

\begin{figure}[t!]
	\centering
	\includegraphics[width=0.9\columnwidth]{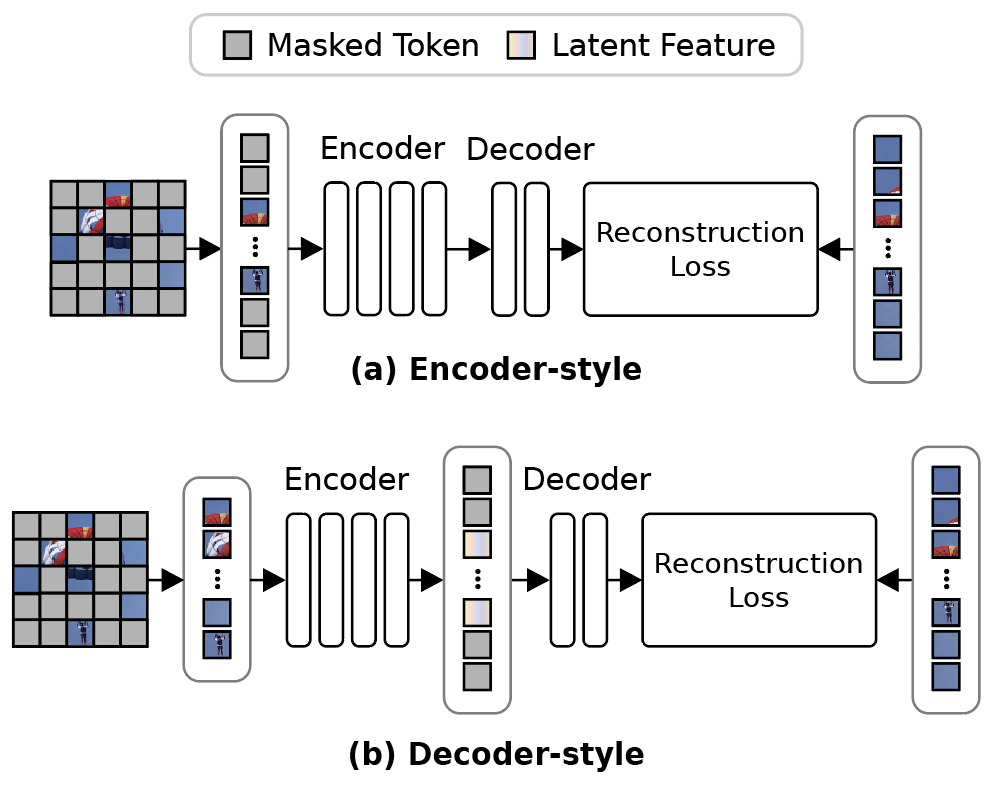}
	\caption{The study of MIM is broadly segmented into two types based on masked token placement: 1) Encoder-style, which reconstructs masked regions within the encoder~\cite{xie2022simmim, bao2021beit, yi2022masked}, and 2) Decoder-style, where reconstruction occurs solely in the decoder~\cite{he2022masked, chen2024context, dong2022bootstrapped}. The recent generative baselines~\cite{wei2023diffusion, zheng2023fast} build on MAE~\cite{he2022masked}, can be seen as decoder-style approaches.}
	\label{fig:encoder_decoder_image}
	\vspace{-0.2cm}
\end{figure}

\Cref{fig:masking_methods} presents the transfer learning performance of each masking method, pre-trained on ImageNet-1K~\cite{imagenet}, and fine-tuned across various recognition tasks and datasets.~\cite{cub, nabirds, inat2017, inat2018, stanfordcars, aircraft, imagenet}
The results indicate that \emph{hybrid masking achieved slightly better performance} across datasets.
We attribute this difference to the fact that diffused masking relies solely on noise. 
Using a random time sampling, when the diffusion noise is weak, the masked image of diffused masking would be very close to the original image, thus rendering the pre-training task trivial---the network will not learn useful semantic representations.
In contrast, hybrid masking would always contain masked portions of the image, forcing the training objective to solve both denoising and de-masking. 
In other words, even when the diffusion part is completely ignored, there is room for the MIM objective alone to still teach the model in hybrid masking.

\subsection{Where Should Noise be Added?}
\label{sec3.3}

Beyond the strategy to combine masking and noising, we look also into where noise/mask should be injected.
The study of MIM is broadly segmented into two competitive approaches, distinguished by the placement of masked tokens, as illustrated in Figure~\ref{fig:encoder_decoder_image}: 1) \textbf{Encoder-style} and 2) \textbf{Decoder-style}. 
The encoder-style~\cite{xie2022simmim, bao2021beit, yi2022masked} in Figure~\ref{fig:encoder_decoder_image} (a), incorporates masked tokens \emph{within the encoder}, with the reconstruction of masked regions occurring across the encoder-decoder structure. In contrast, the decoder-style approach~\cite{he2022masked, chen2024context, dong2022bootstrapped}, illustrated in Figure~\ref{fig:encoder_decoder_image} (b), employs masked tokens solely \emph{within the decoder}, where the primary reconstruction is conducted.

\begin{figure}[t!]
	\centering
	\includegraphics[width=0.6\columnwidth]{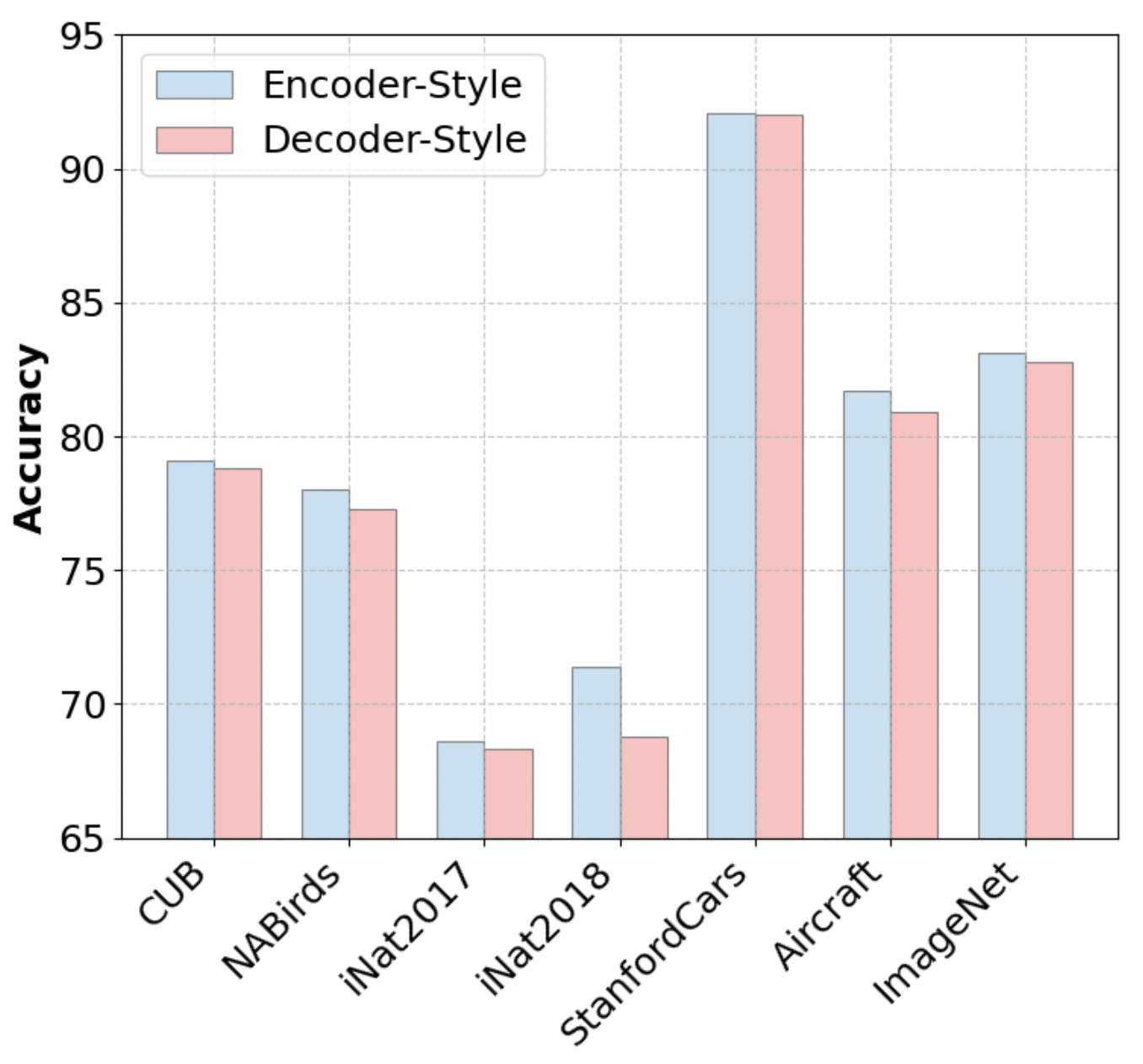}
	\vspace{-1em}
	\caption{We advocate for an encoder-style approach, as it is the encoder that is transferred for downstream fine-tuning. We implemented two naive frameworks of encoder-style and decoder-style generative frameworks. However, the results showed minimal difference, with the encoder-style performing slightly better. \Cref{sec3.4} further explains that a naive implementation of this approach failed to fully leverage its potential.}
	\label{fig:encoder_decoder_style}
	\vspace{-0.2cm}
\end{figure}

A key structural characteristic of the decoder-style is the clear separation between representation learning and reconstruction tasks. The encoder handles representation learning tasks only from visible tokens, while the decoder focuses on the reconstruction tasks using masked tokens.

The recent generative baselines analyzed in \Cref{sec3.2}~\cite{wei2023diffusion, zheng2023fast} reconstruct masked or noisy tokens within the decoder, utilizing MAE~\cite{he2022masked}—a representative decoder-style method of MIM—as their foundation. 
Thus, these approaches can be regarded as decoder-style approaches, but in the context of generative pre-training.

However, we focus on the fact that it is the encoder that is transferred and utilized for downstream fine-tuning. 
Therefore, we advocate for an encoder-style approach, adding corruption and focusing reconstruction within the encoder, hypothesizing that this will allow the benefits of noising which could translate to pre-training.

\begin{figure}[t!]
	\centering
	\includegraphics[width=1\columnwidth]{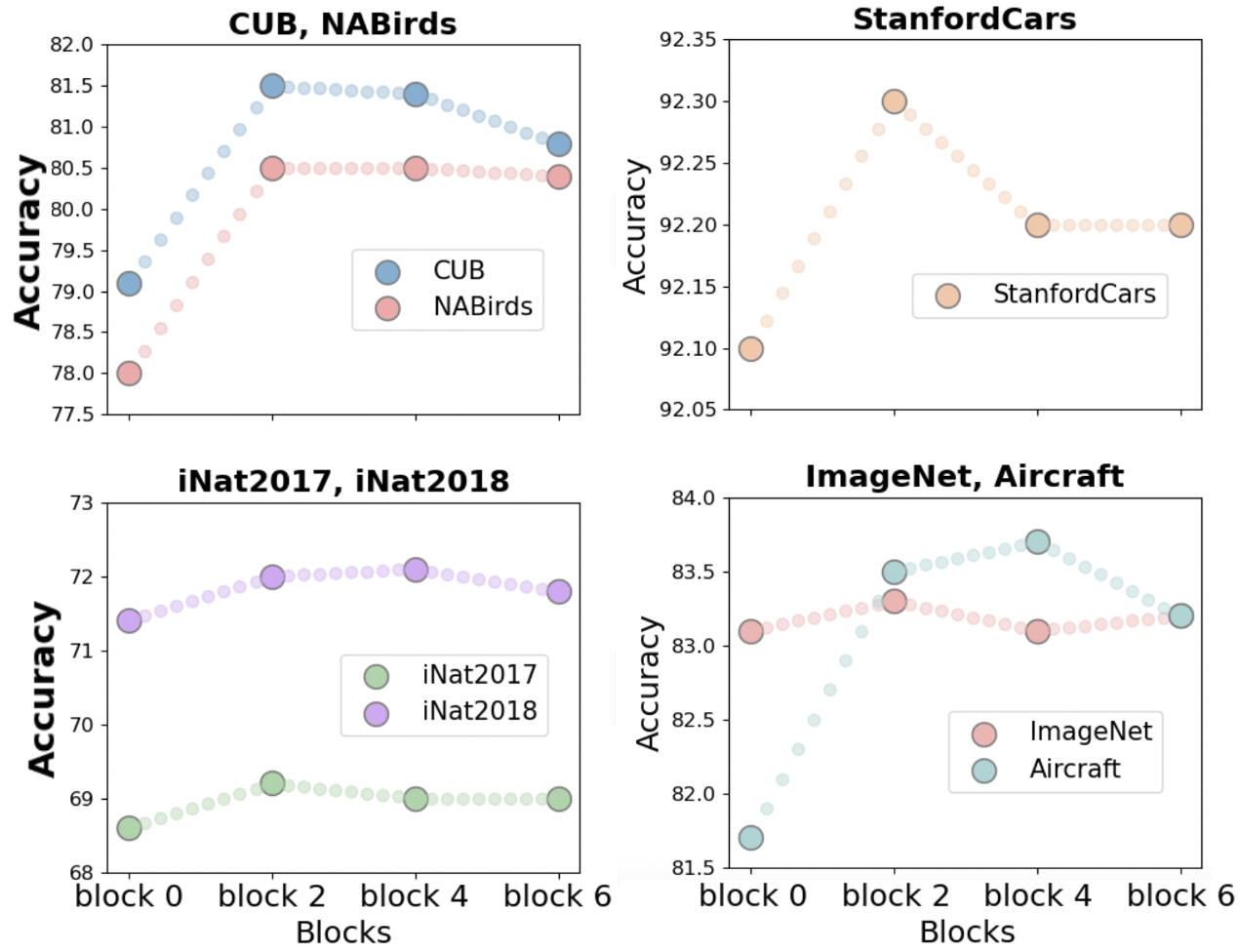}
	\caption{We conducted experiments by introducing noise at various encoder stages (blocks 0, 2, 4, and 6). Results show that adding noise in feature-space (blocks 2, 4, and 6) outperforms pixel-space (block 0) for recognition tasks, with optimal performance at block 2, where high-frequency details are captured.}
	\label{fig:noise_block}
	\vspace{-0.2cm}
\end{figure}

To support this hypothesis, we implemented two naive generative frameworks featuring hybrid masking strategies that differ only in their placement of corruption. We kept all other factors identical.
We then evaluated their performance on transfer learning by pre-training both models on ImageNet-1K~\cite{imagenet} and fine-tuning them on the range of recognition tasks and datasets.

\Cref{fig:encoder_decoder_style} illustrates the transfer learning performance of each masking method. Contrary to our hypothesis, the results revealed minimal performance difference between the two structures. To be specific, in both fine-grained and standard recognition tasks, the difference between the two architectures was minimal, with the encoder-style structure performing only slightly better. 
As we will discuss in \Cref{sec3.4}, this, in fact, is because naively implementing this idea does not show its true potential.


\subsection{Method}
\label{sec3.4}

The encoder-style framework implemented in the above \Cref{sec3.3} is a naive and straightforward approach, as we simply shifted the corruption process to the start of the encoder. 
To properly introduce an encoder-style framework, we analyzed the structural differences between encoder-style and decoder-style approaches and found two key aspects for consideration:
\begin{itemize}
    \item \textbf{Feature-level noise addition}: noise should be added at the feature level; and 
    \item \textbf{Task disentanglement}: the processes of denoising and mask reconstructing should be explicitly disentangled.
\end{itemize}

\begin{figure}[t!]
	\centering
	\includegraphics[width=0.9\columnwidth]{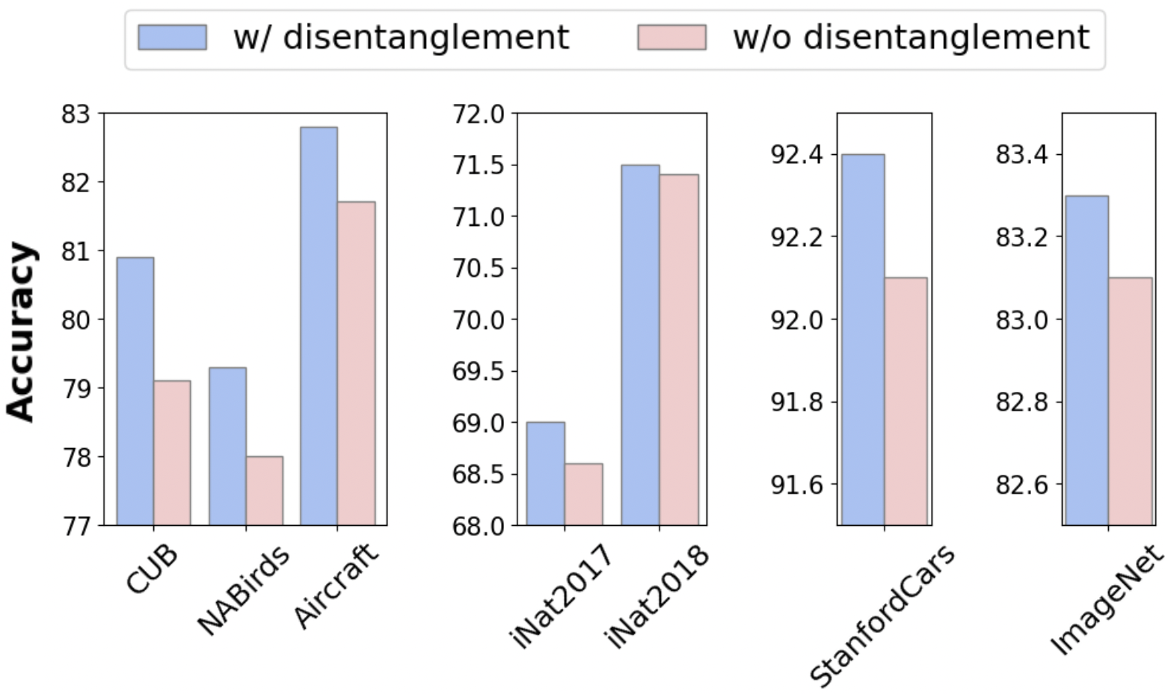}
	\vspace{-1em}
	\caption{We propose an explicit objective to disentangle the de-masking task from the de-noising task, fully harnessing both reconstructions within the encoder. The disruption loss adjusts the weight distribution of the affinity map, minimizing the influence of masked tokens on noisy visible tokens, and enhancing performance across both fine-grained and standard recognition tasks.}
	\label{fig:task_disentanglement}
	\vspace{-0.2cm}
\end{figure}

\paragraph{Feature-level noise addition.}
Much of the success of denoising diffusion models is rooted in the application of noise at the latent (feature) space~\cite{rombach2022high,chen2024deconstructing}.
While pixel-space diffusion exists~\cite{hoogeboom2024simpler}, they need careful strategies on how noise should be applied.
As such, we also suspect this to be the case for pre-training.
However, in the naive implementation that we consider in \Cref{sec3.3}, the decoder-style method adds noise at the latent space immediately before the decoder, whereas the encoder-style approach injects noise in pixel-space, prior to the encoder. 
We thus experimented with adding noise to various blocks within the encoder to investigate the impact of different noise-addition locations.

In \Cref{fig:noise_block}, we conducted experiments by varying the stage at which noise is introduced within the encoder, specifically at different encoder blocks (blocks 0, 2, 4, and 6). 
The transfer learning performance results for recognition tasks verify that adding noise in feature-space (blocks 2, 4, and 6) is more effective than in pixel-space (block 0). 
Additionally, the highest performance observed at `encoder block 2' suggests that noise addition is particularly effective when applied in the lower layers of the encoder, where high-frequency details are captured. 
This result reveals that the injecting noise at the feature level is crucial for maximizing the transfer learning potential of the model.

\paragraph{Task disentanglement.}
Referring to recent studies of MIM~\cite{choi2024emerging, he2022masked, dong2022bootstrapped}, encoder-style approaches often underperform compared to decoder-style approaches. 
This is primarily because masked tokens are trained in a direction orthogonal to that of visible tokens~\cite{choi2024emerging}, which can interfere with the encoding process of visible tokens. 
Because we maintain the structure of hybrid masking, based on the results of \Cref{sec3.2}, we suspect this issue to also arise for generative pre-training with denoising models. 
Specifically, masked tokens may disrupt the encoding process of noisy visible tokens.

The decoder-style approach~\cite{he2022masked, chen2024context, dong2022bootstrapped} avoids this issue by assigning distinct tasks to the encoder and decoder, focusing the encoder on representation learning and the decoder on masked token reconstruction. 
However, as we wish to apply them both to the encoder,
it is essential to explicitly disentangle these two 
within the encoder itself.

\begin{figure*}[t!]
	\centering
	\includegraphics[width=2.1\columnwidth]{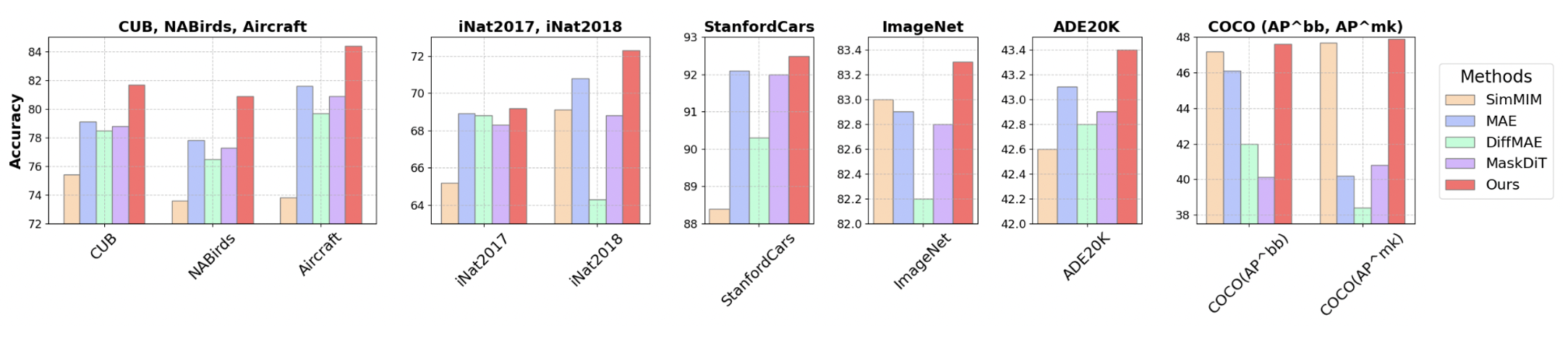}
	\vspace{-2em}
	\caption{The proposed method (Ours) consistently outperforms representative MIM~\cite{xie2022simmim, he2022masked} and generative methods~\cite{wei2023diffusion, zheng2023fast} across a wide range of recognition tasks, capturing diverse frequency details as shown in \cref{fig:kl_divergence} and \cref{fig:qualitative}, that improve accuracy in FGVC, image classification, semantic segmentation, object detection, and instance segmentation tasks.
	}
	\label{fig:main_result}
	\vspace{-0.2cm}
\end{figure*}
\begin{figure*}[t!]
	\centering
	\includegraphics[width=1.7\columnwidth]{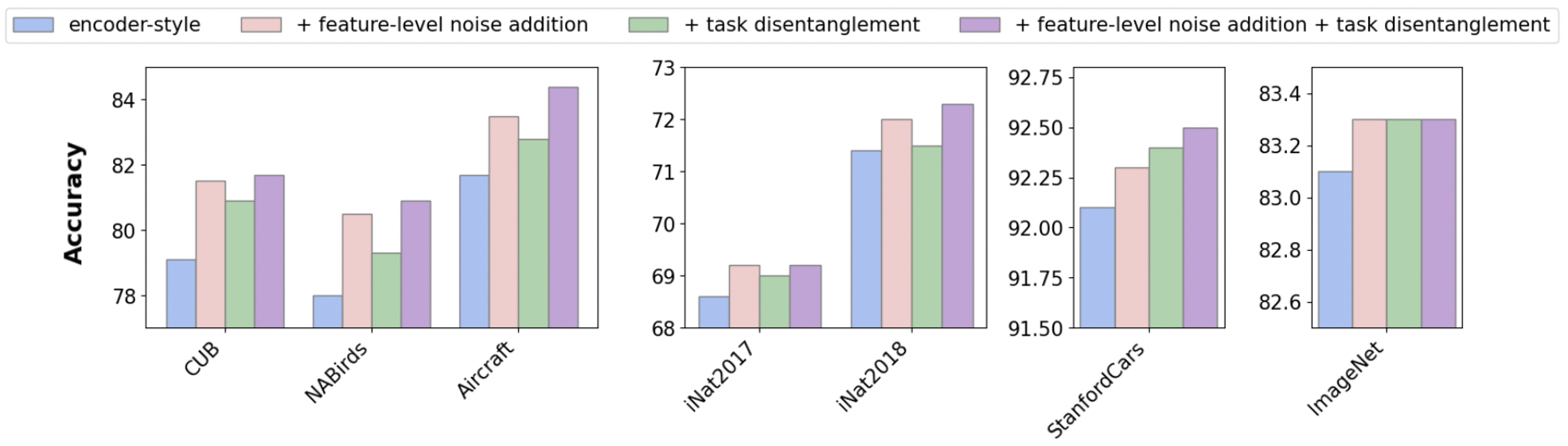}
	\vspace{-1em}
	\caption{We present the ablation study results on the components of the proposed method. Feature-level noise addition and task disentanglement together yield the best results, highlighting the importance of these structural considerations in improving transferability on both fine-grained and standard recognition tasks.
	}
	\label{fig:abl_component}
	\vspace{-0.2cm}
\end{figure*}

To address this, we propose an explicit objective that disentangles the de-masking strategy from the de-noising strategy. 
We introduce disruption loss, a variant of masked token optimization proposed in MTO~\cite{choi2024emerging}, designed to suppress attention between the two different token types. 
Disruption loss leverages per-row sparsities within the affinity matrix, which consists of four quadrants as follows:
\begin{equation}
A = \begin{bmatrix} A_{vv} & A_{vm} \\ A_{mv} & A_{mm} \end{bmatrix}
,
\end{equation}
where $A_{vv}$ represents weights between visible-visible tokens, $A_{vm}$ and $A_{mv}$ represents affinities between visible-masked tokens, and $A_{mm}$ represents weights among masked tokens. The disruption loss $\mathcal{L}_{{d}}$ recalibrates the weight distribution of $A$, minimizing the impact of masked tokens $x_m$ on noisy visible tokens $x_n^t$: 
\begin{equation}
\mathcal{L}_{{d}} = - \sum_{i \in \mathcal{N}} \sum_{j} \tilde{p}_{i,j} \log \tilde{p}_{i,j}
\end{equation}
where $\mathcal{N}$ denotes the index set of noisy visible tokens $x_n^t$, and
\(\tilde{p}\) is an element of $A$, satisfying \(0 < \tilde{p}_{i,j} < 1\) and \(\sum_{j} \tilde{p}_{i,j} = 1\).
The application of $\mathcal{L}_{d}$ reduces interference between different token types and thus ensures effective task disentanglement between denoising and mask reconstruction within the encoder.

The experimental results in \Cref{fig:task_disentanglement} show a performance improvement with this weight adjustment in both fine-grained and standard recognition tasks. Thus, disentangling the de-masking and de-noising strategies through explicit task disentanglement maximizes the transfer potential of the encoder-style approach.

Through these findings, we demonstrate that the encoder-style approach can indeed outperform the decoder-style in generative pre-training frameworks.
\section{Experiments}

\subsection{Implementation Detail}
All experiments in this manuscript were conducted under precisely identical conditions to ensure accurate analysis. To achieve this consistency, we implemented each method, so reported results may differ from original papers.
All experiments used ViT-B~\cite{dosovitskiy2020image} as the backbone architecture applying a unified 400-epoch training schedule on our hardware configuration (4 × A100 GPUs). Pre-training was performed on the ImageNet-1K~\cite{imagenet} classification dataset, followed by fine-tuning on respective downstream task datasets. For transparency, all code, parameters, and detailed experimental settings are available in the Supplementary Material.

\subsection{Main Result}
In \Cref{fig:main_result}, we evaluate the proposed method on diverse tasks, including fine-grained visual categorization (FGVC), image classification, semantic segmentation, object detection, and instance segmentation, each with task-specific datasets. 
FGVC datasets (CUB-200-2011~\cite{cub},  NABirds~\cite{nabirds},  iNaturalist
2017~\cite{inat2017}, iNaturalist 2018~\cite{inat2018}, Stanford Cars~\cite{stanfordcars}, Aircraft~\cite{aircraft}) demand detailed, fine-grained feature learning to distinguish visually similar classes. 
In contrast, standard recognition tasks (ImageNet~\cite{imagenet}), semantic segmentation (ADE20K~\cite{ade20k}), object detection and instance segmentation (COCO~\cite{coco}) emphasize broader spatial
details at different levels of granularity.

The proposed method (Ours) consistently outperforms representative MIM~\cite{xie2022simmim, he2022masked} and generative methods~\cite{wei2023diffusion, zheng2023fast} across tasks. In FGVC tasks, our method effectively captures high-frequency, localized features, as shown in \Cref{fig:kl_divergence} and \Cref{fig:qualitative}, surpassing comparison methods in accuracy. Even in standard recognition tasks, where spatial detail is key, our method shows favourable gains, highlighting that task disentanglement and feature-space noise injection enhance transfer potential and capture diverse frequency information, as demonstrated in \Cref{fig:kl_divergence}.

\subsection{Ablation on Components}
Figure~\ref{fig:abl_component} shows the ablation study results on our proposed method, progressively enhanced by feature-level noise addition and task disentanglement, individually and in combination, across various recognition benchmarks.

Feature-level noise consistently improves accuracy in both fine-grained (CUB, NABirds, Aircraft, iNat2017, iNat2018, StanfordCars) and standard recognition tasks (ImageNet). Task disentanglement further enhances performance by separating de-masking and de-noising processes. The combination of both techniques yields the best results. This result underscores the importance of structural design considerations for optimal transfer learning. Further ablation studies are available in the Supplementary Material.

\section{Conclusion}

We have analyzed why current noising-based pre-training struggles with recognition tasks and provided architectural guidelines. 
By applying our findings, we achieve notable improvements over existing MIM and other generative approaches.

\paragraph{Limitations and future work.} 
Our analysis is currently limited to recognition tasks.
It would thus be interesting to broaden the scope of our study to other applications.

{
    \small
    \bibliographystyle{ieeenat_fullname}
    \bibliography{main}
}


\end{document}